\documentclass[10pt,twocolumn,letterpaper]{article}

\usepackage{cvpr}              % To produce the CAMERA-READY version
\usepackage{color,soul}

\usepackage[dvipsnames]{xcolor}

\usepackage{multirow}
\usepackage{booktabs}
\usepackage{algorithm}
\usepackage{algpseudocode}
\definecolor{cvprblue}{rgb}{0.21,0.49,0.74}
\usepackage[pagebackref,breaklinks,colorlinks,citecolor=cvprblue]{hyperref}

\def\paperID{4002} % *** Enter the Paper ID here
\def\confName{CVPR}
\def\confYear{2024}

\title{DiSR-NeRF: Diffusion-Guided View-Consistent Super-Resolution NeRF}
\author{Jie Long Lee  \qquad  Chen Li \qquad Gim Hee Lee\\
Department of Computer Science, National University of Singapore\\
{\tt\small ljielong@comp.nus.edu.sg, lichen@u.nus.edu, gimhee.lee@nus.edu.sg}
}

\begin{document}
\maketitle
\begin{abstract}
We present DiSR-NeRF, a diffusion-guided framework for view-consistent 
super-resolution (SR) NeRF. Unlike prior works, we circumvent the requirement for high-resolution (HR) reference images by leveraging existing powerful 2D super-resolution models.
Nonetheless, independent SR 2D images are often inconsistent across different views. We thus propose Iterative 3D Synchronization (I3DS) to mitigate the inconsistency problem via the inherent multi-view consistency property of NeRF. Specifically, our I3DS alternates between upscaling low-resolution (LR) rendered images with diffusion models, and updating the underlying 3D representation with standard NeRF training. We further introduce Renoised Score Distillation (RSD), a novel score-distillation objective for 2D image resolution. Our RSD combines features from ancestral sampling and Score Distillation Sampling (SDS) to generate sharp images that are also LR-consistent. Qualitative and quantitative results on both synthetic and real-world datasets demonstrate that our DiSR-NeRF 
can achieve better results on NeRF super-resolution compared with existing works. Code and video results available at the project website\footnote[1]{\hyperlink{https://github.com/leejielong/DiSR-NeRF}{https://github.com/leejielong/DiSR-NeRF}}.
\end{abstract}    
\section{Introduction}
\label{sec:intro}

\begin{figure}[t]
\begin{center}
\centering

\includegraphics[width=\linewidth, trim=0 50 0 0]{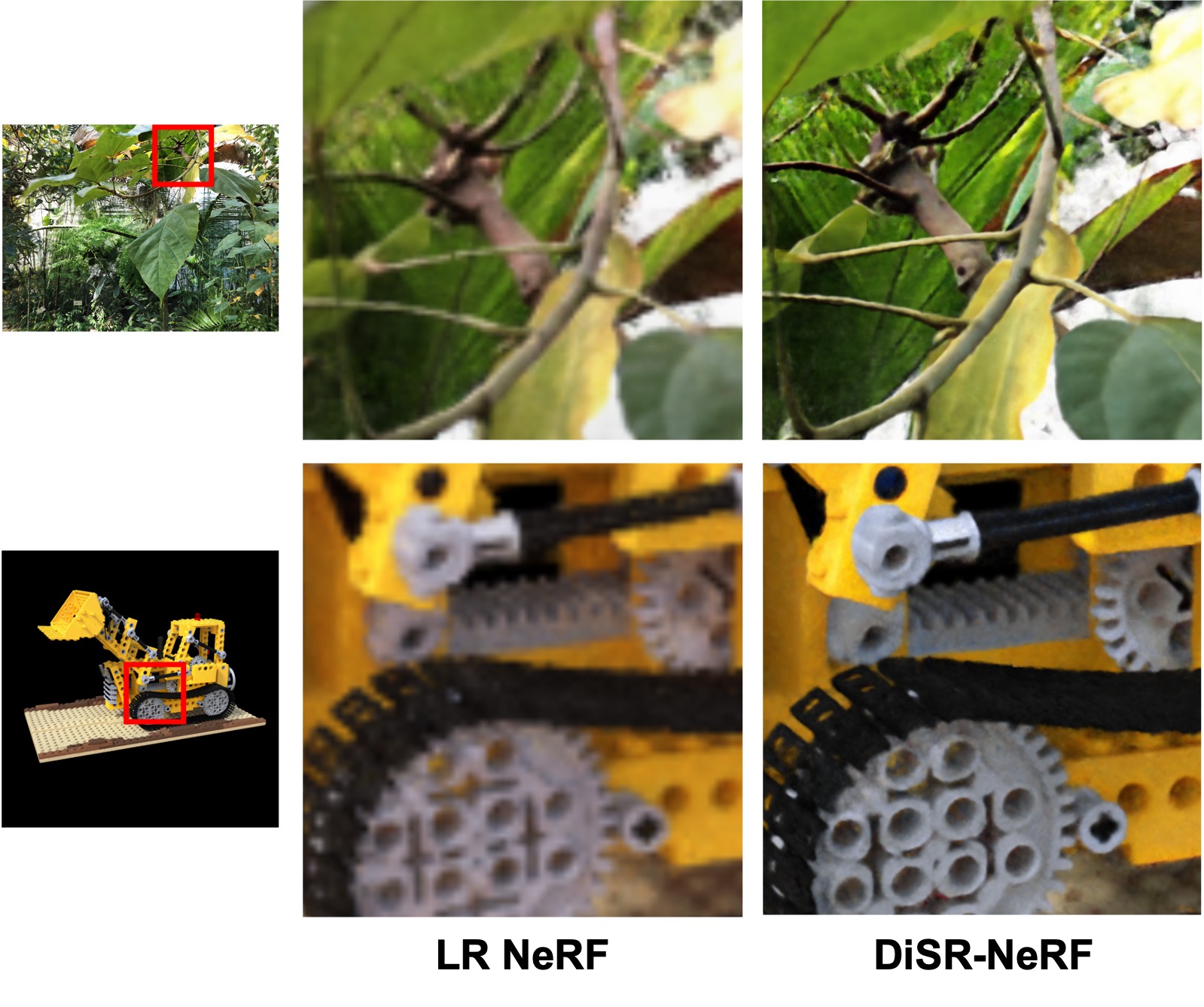}

\end{center}
\caption{
Our DiSR-NeRF distils super resolution priors from a 2D diffusion upscaler to generate high quality details from low resolution NeRFs.
}
\vspace{-0.5cm}
\label{fig:teaser}
\end{figure}

Novel view synthesis is a long-standing problem in computer vision with significant real-world applications. Recently, neural radiance fields (NeRFs) have emerged as a powerful representation, achieving state-of-the-art performance in novel view synthesis.

Since the pioneering work on NeRFs by \cite{mildenhall2020nerf}, many follow up works have explored the improvement of NeRF's speed \cite{sun2021direct,muller2022instant,fridovich2022plenoxels,chen2022tensorf}, 
fidelity \cite{barron2021mipnerf,barron2022mipnerf360,wang20234knerf}, scale 
\cite{mi2023switchnerf,tancik2022blocknerf}, robustness \cite{lin2021barf,chen2022local,SCNeRF2021}, and generalizability \cite{Yu20arxiv_pixelNeRF,wang2021ibrnet,kosiorek2021nerfvae,chen2021mvsnerf,mine,xu2022point,xu2022sinnerf}. 
However, one important aspect of NeRFs that has not been well explored is super-resolution. In real-world scenarios, imaging devices may be limited in resolution (\ie, drones, CCTVs, etc.) and consequently high-resolution multi-view images may be unavailable. With low-resolution inputs, 
NeRF struggles to represent the high-quality details of the underlying 3D scenes.
In this work, we tackle the task of NeRF super-resolution, which aims to learn high-resolution implicit representation of 3D scenes from only low-resolution images. One possible direction is to design a generative 3D super-resolution model, which however, requires large datasets of high-resolution multi-view images for training.Collecting such large-scale, high-resolution multi-view data is labor-intensive and requires expensive equipment to obtain accurate scans.
On the other hand, large 2D high-resolution image datasets such as LAION-5B \cite{schuhmann2022laion5b} are publicly available and have been used to train powerful 2D super-resolution models. We thus propose to leverage knowledge from the 2D super-resolution models to circumvent the requirements for HR images.

Naively upscaling individual LR training images with 2D super-resolution methods produces SR images that may not be consistent across views. A NeRF trained on such images produces blurred details as SR details may not agree across different camera views. Recent works such as Super-NeRF \cite{han2023supernerf} resolve this by searching the latent space of a 2D upscaler for view-consistent SR results, however the framework only enforces low-resolution view-consistency.
In this work, we propose DiSR-NeRF, a Diffusion-guided Super-Resolution NeRF method which produces NeRFs with high resolution and view-consistent details. 

Our method comprises two key components.
1) We propose the two-stage \textbf{Iterative 3D Synchronization (I3DS)} to solve the cross-view inconsistency problem. 
We first refine the rendered images from the LR NeRF with a diffusion-based 2D super-resolution model, and subsequently synchronize the details into 3D through standard NeRF training. The alternating process between the two stages guides the NeRF to converge to view-consistent details.
2) We introduce the \textbf{Renoised Score Distillation (RSD)} objective to get the best from both worlds of ancestral sampling and Score Distillation Sampling (SDS).  
Particularly, we observe that the default ancestral sampling used in diffusion-based super resolution can generate details that are structurally inconsistent with the conditioned LR image (\cf \cref{fig:2d} in the experiments). This may aggravate the inconsistency across different views. On the other hand, the Score Distillation Sampling (SDS) objective commonly used in Text-to-3D generation can produce LR-consistent features but with limited details. 
The optimization target of RSD is designed as the intermediate denoised latents of the ancestral sampling trajectory. This transforms the ancestral sampling process into an optimization framework, generating coarse-to-fine details over the course of optimization. As a result, RSD is able to achieve sharper details compared to SDS while also producing LR-consistent features compared to ancestral sampling.

Our method requires only low-resolution multi-view images of a target scene, thus alleviating the cumbersome need for high-resolution reference images or large scale multi-view HR image datasets. Our qualitative and quantitative results show that DiSR-NeRF can outperform existing baselines to achieve effective super-resolution NeRF. Our contributions are as follows:
\begin{itemize}
  \item We introduce DiSR-NeRF, a method that achieves high quality super-resolution NeRF using only LR training images and a pretrained 2D diffusion upscaler. 
  \item We propose Iterative 3D Synchronization (I3DS) 
  that achieves convergence to view-consistent SR details.
  \item We design Renoised Score Distillation (RSD), a score-distillation objective that produces sharper details and maintains consistency towards the conditioned LR image.
\end{itemize}
\section{Related Works}
\label{sec:related}

\subsection{2D Image Super-Resolution}
Image super-resolution is inherently an ill-posed problem since there can be a distribution of possible SR solutions for a LR image. Most state-of-the-art approaches therefore seek to learn the conditional distribution $p(\mathbf{x} \mid \mathbf{y}, \mathbf{w})$ %\boldsymbol\theta)$ 
of possible SR images $\mathbf{x}$ that correspond to the conditioned LR image $\mathbf{y}$ using a generative model parameterized by $\mathbf{w}$. 
\vspace{-6mm}
\paragraph{GANs.} Generative Adversarial Networks (GANs) are a class of generative models that have demonstrated impressive results in image SR. GAN-based SR models \cite{ledig2017photorealistic,wang2018esrgan,wang2021realesrgan} utilize adversarial training objectives to produce SR images that appear photorealistic to human visual perception. 
However, GAN-based SR models are often difficult to train and may encounter mode collapse. 
\vspace{-6mm}
\paragraph{Normalizing Flows.} Flow-based SR models \cite{lugmayr2020srflow,jo2021srflowda,yao2023local,liang21hierarchical} employ conditional normalizing flows to model the conditional SR image distribution. Unlike GANs, flow-based SR models learn to explicitly compute the probability density of SR images conditioned on the LR image. However, flow-based SR models are restricted to invertible architectures due to bijectivity constraints and lack expressiveness compared to other approaches.
\vspace{-6mm}
\paragraph{Diffusion Models.} 
Diffusion-based SR models \cite{saharia2021image,ho2021cascaded,rombach2022highresolution,li2021srdiff,xia2023diffir} learn to generate SR images from pure Gaussian noise using a trained denoising network to iteratively restore structure and details while being conditioned on a text prompt and LR image. In our work, we use the Stable Diffusion $\times$4 Upscaler (SD$\times$4), which is a pretrained latent diffusion upscaler \cite{rombach2022highresolution} to guide the generation of high-resolution details in 3D. Latent diffusion models use a pretrained variational autoencoder (VAE) to project images to a lower-dimensional latent space for diffusion. Executing the diffusion process in the latent space improves speed and GPU memory usage. 

\subsection{Super-Resolution NeRF}
Super-Resolution NeRF is currently an area that has yet to be well-explored. Some works improve NeRF details via anti-aliasing or ray supersampling \cite{barron2021mipnerf,barron2022mipnerf360,wang2021nerf-sr}, but remain fundamentally limited by the level of detail available in the input images. Other works achieve super-resolution NeRF \cite{wang2021nerf-sr,refsr} under a reference-guided setting, requiring HR reference images of the target scene to be available. Such requirements can be impractical when only low resolution imaging solutions are available.

Super-NeRF \cite{han2023supernerf} is a recent work with similar motivation to ours, and it achieves high-resolution detail generation for LR NeRF by searching the latent space of ESRGAN \cite{wang2018esrgan} for view-consistent solutions. However, the proposed framework only explicitly constraints view consistency in the LR domain. In contrast to Super-NeRF, we utilize the score function of diffusion-based upscaler models to produce sharp and coherent SR details. Furthermore, we also introduce a 3D synchronization mechanism to converge on view-consistent features.
\begin{figure}[t]
\begin{center}
\centering

\includegraphics[width=\linewidth]{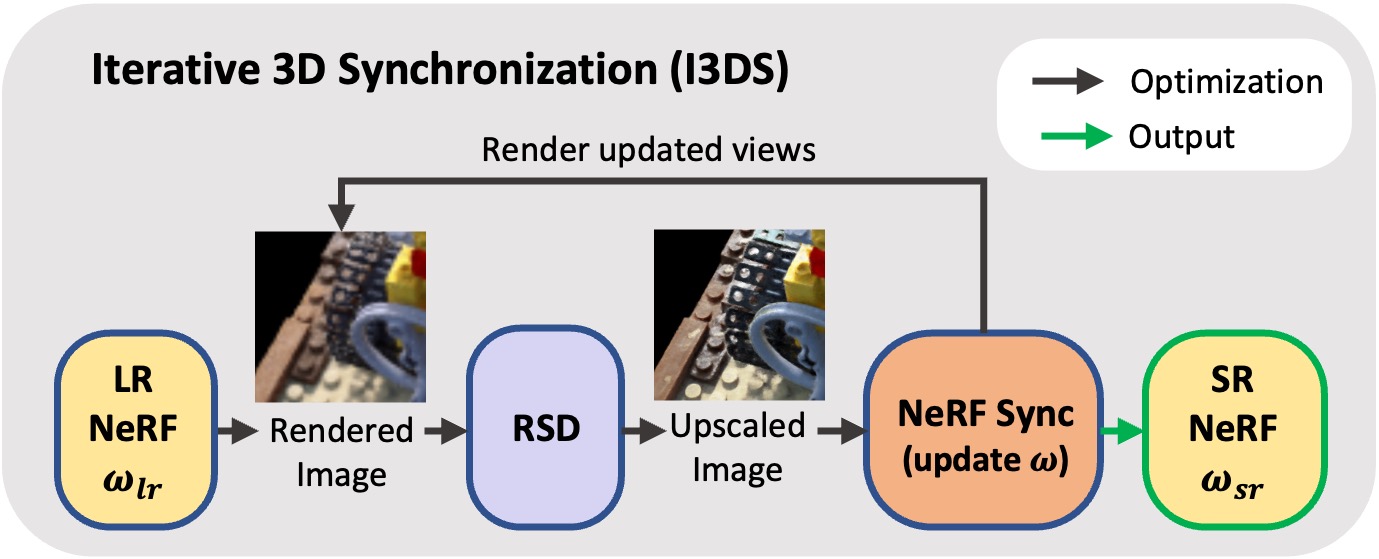}

\end{center}
\caption{
I3DS seperates upscaling and NeRF fitting in seperate, alternate stages. NeRF renders are upscaled via RSD in the upscaling stage, and upscaled images are used as training images to learn view-consistent details. The two stage process is repeated over several cycles to achieve detail convergence.}
\vspace{-0.cm}
\label{fig:i3ds}
\end{figure}
\begin{figure*}[t]
\begin{center}
\centering

\includegraphics[width=\linewidth, trim=0 20 0 0]{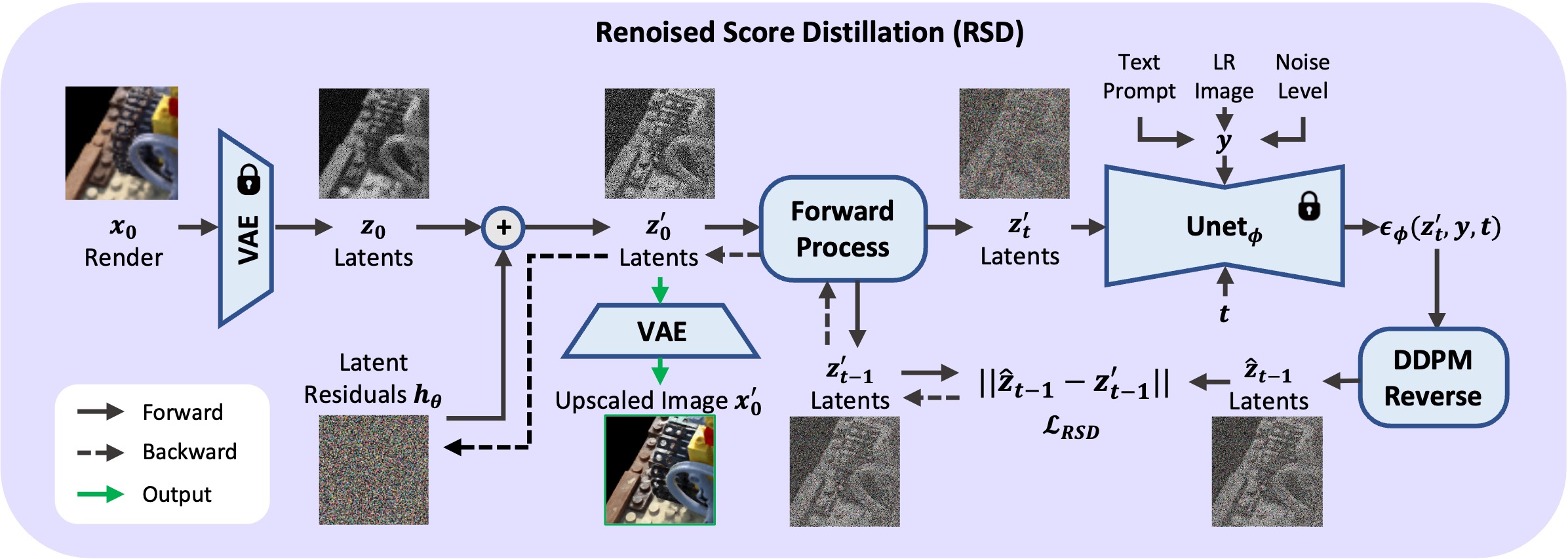}

\end{center}
\caption{
Our RSD produces LR-consistent HR details by optimizing $\mathbf{z}'_{t-1}$ towards predicted denoised latents $\hat{\mathbf{z}}_{t-1}$ following a linearly decreasing time schedule. After optimization, the residuals $\mathbf{h}_\theta$ contain HR details that is added to $\mathbf{z}_0$ to obtain upscaled latents $\mathbf{z}'_0$, which is decoded into LR-consistent upscaled images $\mathbf{x}'_0$. Refer to the text in Sec.~\ref{subsection:RSD} for more details.}
\vspace{-0.4cm}
\label{fig:rsd}
\end{figure*}
\section{Our Method}
\label{sec:method}

\subsection{Preliminaries}
\label{subsection:prelim}
\paragraph{Diffusion Models.} 
Diffusion models \cite{ho2020denoising,nichol2021improved,song2020generative,song2021scorebased,song2022denoising} are generative models which transform a sample from a noise distribution towards a data distribution using a fixed forward process and a learned reverse process. The forward process introduces noise to data based on a predetermined noising schedule and gradually destroys detail and structure. A data sample $\mathbf{z}_0$ can be noised into $\mathbf{z}_t$ using the closed-form formula \cite{ho2020denoising} of the forward process: 
\begin{equation}
\label{eqn:forward}
\mathbf{z}_t=\sqrt{\bar{\alpha_t}}\mathbf{z}_0+\sqrt{1-\bar{\alpha_t}}{\boldsymbol\epsilon}
\end{equation}
which produces noised latents $\mathbf{z}_t$ at timestep $t\in[0,T]$ given timestep-dependent noising coefficient $\bar{\alpha}_t$ and Gaussian noise sample $\boldsymbol\epsilon\in\mathcal{N}(0,\mathbf{I})$. 

The reverse process restores structure from noise. It is learned by a network $\boldsymbol\phi$ that is trained to denoise a noised latent $\mathbf{z}_t$ by predicting its noise component $\boldsymbol\epsilon_\phi(\mathbf{z}_t,y,t)$ conditioned upon $y$ (text prompt/image) and timestep $t$. In diffusion image generators, $y$ comprises the embedded text prompt, while in diffusion upscalers, a lightly noised LR image and the image noising level is also included. Diffusion models are typically trained with an evidence lower bound (ELBO) objective \cite{ho2020denoising} given by:
\begin{equation}
    \mathcal{L}=\mathbb{E}_{t\sim U(0,T),\epsilon\sim\mathcal{N}(\mathbf{0},\mathbf{I})}[\gamma(t)\|\boldsymbol\epsilon_\phi(\mathbf{z}_t,y,t)-\boldsymbol\epsilon\|^2_2].
\end{equation}
The sampling process which is also referred to as ancestral sampling, iteratively denoises a Gaussian noise sample $\mathbf{z}_T$ according to: 
\begin{equation}
\label{eqn:reverse}
    \mathbf{z}_{t-1} = \frac{1}{\sqrt{\alpha_t}}\Bigl(\mathbf{z}_t-\frac{1-\alpha_t}{\sqrt{1-\bar{\alpha_t}}}\boldsymbol\epsilon_\phi(\mathbf{z}_t,y,t)\Bigr)+\sigma_t\boldsymbol\epsilon
\end{equation}
for a DDPM \cite{ho2020denoising} scheduler, where $\sigma_t$ is the standard deviation of Gaussian noise samples. In ancestral sampling, the latent $\mathbf{z}_t$ is gradually denoised into $\mathbf{z}_0$ from the data distribution via a sample trajectory given by:
\begin{equation}
    (\mathbf{z}_T,\mathbf{z}_{T-1},...,\mathbf{z}_1,\mathbf{z}_0),
\end{equation}
where $\mathbf{z}_0$ is the generated output image.

\vspace{-3mm}
\paragraph{Score Distillation Sampling.} Diffusion models are also score-based generative models \cite{song2021scorebased,wang2022score} which learn a score function $\nabla_\mathbf{z}\text{log}p_t(\mathbf{z})$ as the gradient of the log probability density with respect to data. 
Score Distillation Sampling (SDS) \cite{poole2022dreamfusion,wang2022score} is an optimization objective that uses the score function of pretrained diffusion models to provide gradients that guide the optimization of a differentiable image parametrization \cite{mordvintsev2018differentiable} $\mathbf{z}=g(\theta)$ towards the mode of a conditional probability distribution $p(\mathbf{z}_t \mid y)$. SDS has demonstrated remarkable results in Text-to-3D generation \cite{poole2022dreamfusion,wang2022score,lin2023magic3d,fantasia,wang2023prolificdreamer,metzer2022latent,liu2023zero1to3,huang2023dreamtime,zhu2023hifa}, but is known to encounter issues such as over-saturation and over-smoothing \cite{wang2023prolificdreamer}.

Given an image $\mathbf{z}_0=g(\theta)$, a pretrained diffusion model predicts the noise component $\boldsymbol\epsilon(\mathbf{z}_t,y,t)$ which is related to the score function $\nabla_{\mathbf{z}_t}\text{log} p(\mathbf{z}_t \mid y)$ by: 
\begin{equation}
\boldsymbol\epsilon(\mathbf{z}_t,y,t)=-\sigma_t\nabla_{\mathbf{z}_t}\text{log} p(\mathbf{z}_t \mid y).
\end{equation}
Consequently, \cite{poole2022dreamfusion} proposes the SDS objective as:
\begin{equation}
\nabla_{\theta}\mathcal{L}_{SDS}=\mathbb{E}_{t,\epsilon}[\gamma(t)(\boldsymbol\epsilon_\phi(\mathbf{z}_t,y,t)-\boldsymbol\epsilon)\frac{\partial{\mathbf{z}_0}}{\partial{\theta}}].
\end{equation}
In 3D, $\mathbf{z}_0=g(\theta)$ is a NeRF render where $g(\cdot)$ corresponds to the volume rendering function and $\theta$ represents the NeRF parameters. In 2D, $g(\cdot)$ is an identity transform and $\theta$ is the pixel values of the image under optimization. In this paper, we use the 2D formulation of $g(\theta)$.

\subsection{Iterative 3D Synchronization (I3DS)}
\label{subsection:i3ds}
In our initial experiments, we observed that directly applying SDS on NeRF renders produces blurred details and would fail to converge on detailed reconstructions. 
We postulate that this is because SDS supervision is only provided over a small
local patch of rays in each training step, which does not optimize NeRF towards globally consistent features. Rendering multiple patches concurrently is also intractable due to significant GPU memory required.
To resolve this issue, we propose Iterative 3D Synchronization (I3DS)
which disentangles upscaling and NeRF synchronization by performing both processses in seperate
alternate stages. 

\cref{fig:i3ds} illustrates our I3DS framework. The first stage is the \textbf{upscaling-stage}. Starting from a NeRF $\boldsymbol\omega_{lr}$ pretrained from low resolution inputs, we render images in 4$\times$ resolution from all training poses. Each rendered image is then independently upscaled using RSD (refer to \cref{subsection:RSD}) with SD$\times$4 to generate high resolution details. 
Since the initial rendered images are less detailed, the upscaling process can generate varying HR details which may not be multi-view consistent. 
The second stage is the \textbf{synchronization-stage}. This stage resolves multi-view inconsistency by using the RSD-refined images as training inputs for the NeRF $\boldsymbol\omega$ that is initialized from $\boldsymbol\omega_{lr}$. During synchronization, NeRF $\boldsymbol\omega$ is updated using the standard NeRF training procedure \cite{mildenhall2020nerf} where rays are randomly sampled across all training views. Here, we leverage the view-consistent property of NeRF to capture coherent details across views. This stage transfers view-aligned details generated in the upscaling stage into the 3D representation.

There is a synergistic effect between the upscaling and synchronization stages. The synchronization-stage enables NeRF to capture view-consistent details and inconsistent details are naturally discarded. Consequently, the upscaling-stage receives increasingly detailed and view-consistent input images rendered from NeRF. This allows RSD to generate additional details with lower cross-view inconsistency. Furthermore, since RSD is always conditioned on the original LR training images, the I3DS process does not degrade into degenerate solutions. Over successive iterations, I3DS updates NeRF $\boldsymbol\omega$ to learn highly detailed and view-consistent features to produce SR NeRF $\boldsymbol\omega_{sr}$.

Our I3DS process shares some similarity to NeRF editing approaches \cite{instructnerf2023,nguyenphuoc2022snerf}
which also utilize an alternating training regime to exploit the multi-view consistent property of NeRFs. Nonetheless, we differ from \cite{instructnerf2023,nguyenphuoc2022snerf} by performing optimization processes in both stages. Furthermore, we optimize all renders concurrently and replace all training images in a single batch after RSD optimization
to achieve efficient parallelization and reduce unnecessary image-to-latent encoding. This design consideration achieves 4$\times$ reduction in optimization duration.

\subsection{Renoised Score Distillation (RSD)}
\label{subsection:RSD}
The upscaling-stage in I3DS involves upscaling NeRF renders in 2D. A straightforward solution would be to use ancestral sampling, which is the de-facto method for diffusion-based upscalers. However, we find that using ancestral sampling with I3DS does not lead to high quality results. Although ancestral sampling produces sharp SR images, structural features may deviate from the LR image conditioning which aggravates cross-view inconsistencies (\cf \cref{fig:2d} for visualization).
Another approach would be to use SDS optimization for 2D upscaling in lieu of ancestral sampling. 
However, SDS produces images that are less detailed than ancestral sampling despite being LR consistent. The over-smoothed results of SDS optimization, which has also been observed in \cite{wang2023prolificdreamer}, limits I3DS from producing high quality NeRFs. 
These observations lead us to the idea of getting the best of both worlds -- we propose Renoised Score Distillation (RSD) to incorporate elements of SDS optimization into the ancestral sampling process to achieve detailed SR images that are also LR consistent. 

As discussed in \cref{subsection:prelim}, ancestral sampling follows a latent trajectory $(\mathbf{z}_T,\mathbf{z}_{T-1},...,\mathbf{z}_1,\mathbf{z}_0)$ that gradually transforms a Gaussian noise sample $\mathbf{z}_T$ to a data sample $\mathbf{z}_0$. Given a noisy latent $\mathbf{z}_t$ obtained from encoding and noising a source image $\mathbf{x}_0$, we set our optimization target as the previous timestep latents $\mathbf{z}_{t-1}$ of the ancestral sampling trajectory. Additionally, we use a linearly decreasing time schedule that follows ancestral sampling instead of randomly sampled timesteps $t$ as in SDS.
As a result, we incrementally build details onto our optimized image similar to ancestral sampling. We find that our RSD is able to achieve sharper details compared to SDS and with improved LR consistency compared to ancestral sampling. In contrast to ancestral sampling that uses one noise prediction $\boldsymbol\epsilon_\phi(\mathbf{z}_t,y,t)$ to obtain $\mathbf{z}_{t-1}$ from $\mathbf{z}_t$, our RSD guides the image under optimization (parametrized by latent residuals $\theta$) towards $\mathbf{z}_{t-1}$ over multiple noise predictions. Our optimization objective is given by:
\begin{equation}
\label{eqn:rsd}
\mathcal{L}_{RSD}=\|\mathbf{z}_{t-1}-\hat{\mathbf{z}}_{t-1}\|, 
\end{equation}
where $\mathbf{z}_{t-1}$ is the current noised latent at $t-1$, and $\hat{\mathbf{z}}_{t-1}$ is the predicted denoised latent from $\mathbf{z}_t$.
Unlike SDS which is defined as a loss gradient that is applied in $\mathbf{z}_0$ space, RSD is formulated as a loss function in order to backpropagate gradients through $\mathbf{z}'_{t-1}$.

\cref{fig:rsd} provides a detailed illustration of the RSD optimization process. We first interpolate a randomly sampled image patch $\mathbf{x}_0$ by 4$\times$ and encode it to latent $\mathbf{z}_0$ using the pretrained VAE encoder of SD$\times$4. We then create zero-initialized learnable latent residuals $\mathbf{h}_\theta$ for each $\mathbf{z}_0$ such that
\begin{equation}
\mathbf{z}'_0 = \mathbf{z}_0 + \mathbf{h}_\theta,
\end{equation}
where $\mathbf{z}'_0$ is the refined latents representing the upscaled image $\mathbf{x}'_0$. Subsequently, we apply the forward process in \cref{eqn:forward} twice
on $\mathbf{z}'_0$ at timesteps $t$ and $t-1$ to obtain two noised latents $\mathbf{z}'_t$ and $\mathbf{z}'_{t-1}$. The UNet backbone of SD$\times$4 (parameterized by $\phi$) then takes $\mathbf{z}'_t$ and conditioning $y$ (comprising text prompt, LR image, noise level) as input to predict noise residual $\boldsymbol\epsilon_{\phi}(\mathbf{z}'_t,y,t)$. 

We then pass $\boldsymbol\epsilon_{\phi}(\mathbf{z}'_t,y,t)$ to \cref{eqn:reverse} of the DDPM reverse process to construct the predicted denoised latent $\hat{\mathbf{z}}_{t-1}$. Finally, we compute the L1 error between $\mathbf{z}'_{t-1}$ and $\hat{\mathbf{z}}_{t-1}$ using \cref{eqn:rsd} and backpropagate gradients through $\mathbf{z}'_{t-1}$ towards $\mathbf{h}_\theta$. After optimization, $\mathbf{h}_\theta$ contains the latent HR residuals that can added to $\mathbf{z}_0$ to produce SR latents $\mathbf{z}'_0$. We can then decode $\mathbf{z}'_0$ using the VAE decoder of SD$\times$4 to obtain SR image $\mathbf{x}'_0$ that is used for training in the NeRF synchronization stage of I3DS. In the supplementary, we provide the pseudocode for I3DS and RSD. We also relate the formulation of RSD to SDS and we show that RSD can be viewed as a \textit{renoised} variant of SDS.

\begin{figure*}[t]
\begin{center}
\centering

\includegraphics[width=\linewidth, trim=0 20 0 0]{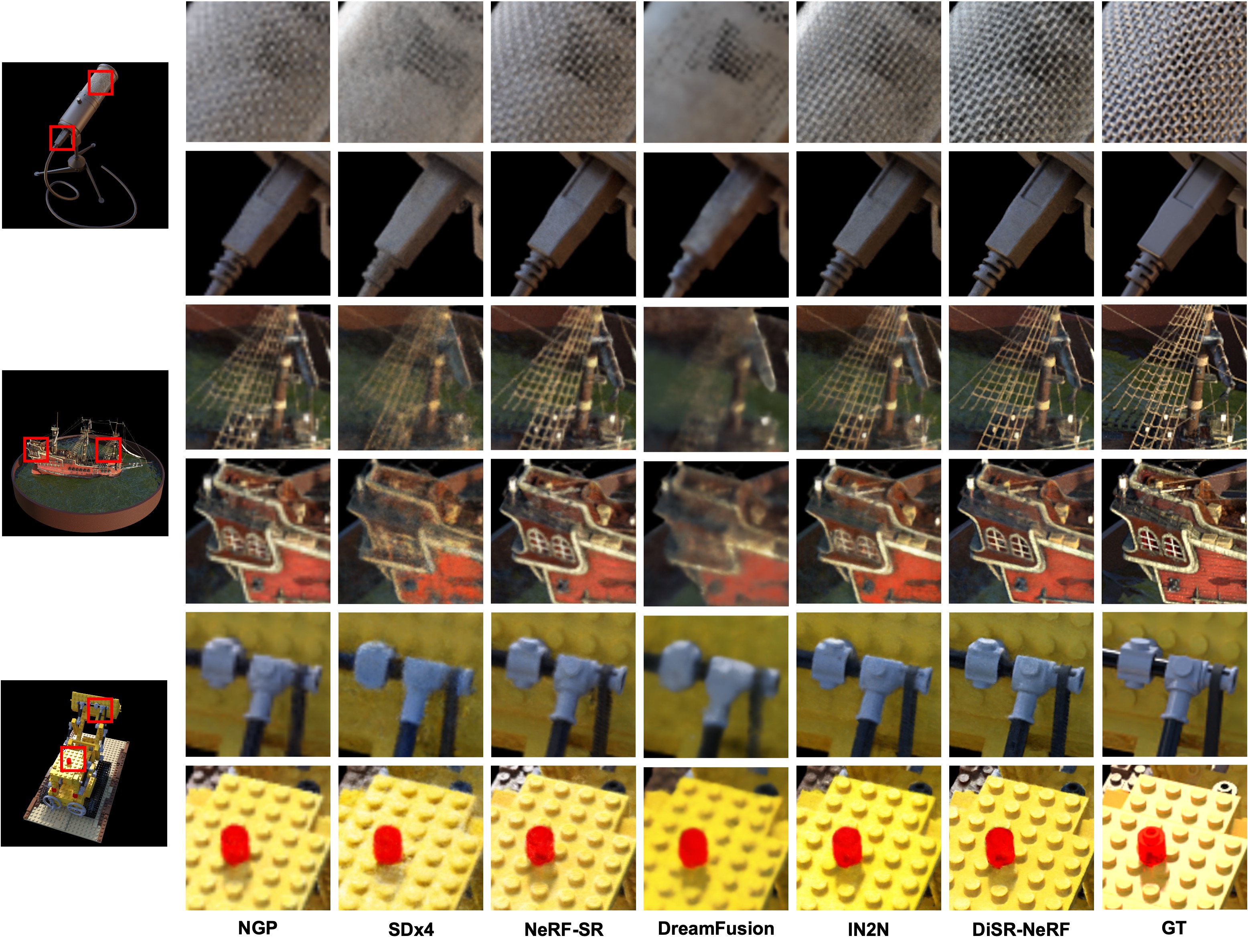}

\end{center}
\caption{
Qualitative Results on NeRF-Synthetic Dataset.}
\vspace{-0.cm}
\label{fig:blender}
\end{figure*}
\begin{figure*}[t]
\begin{center}
\centering

\includegraphics[width=\linewidth, trim=0 20 0 0]{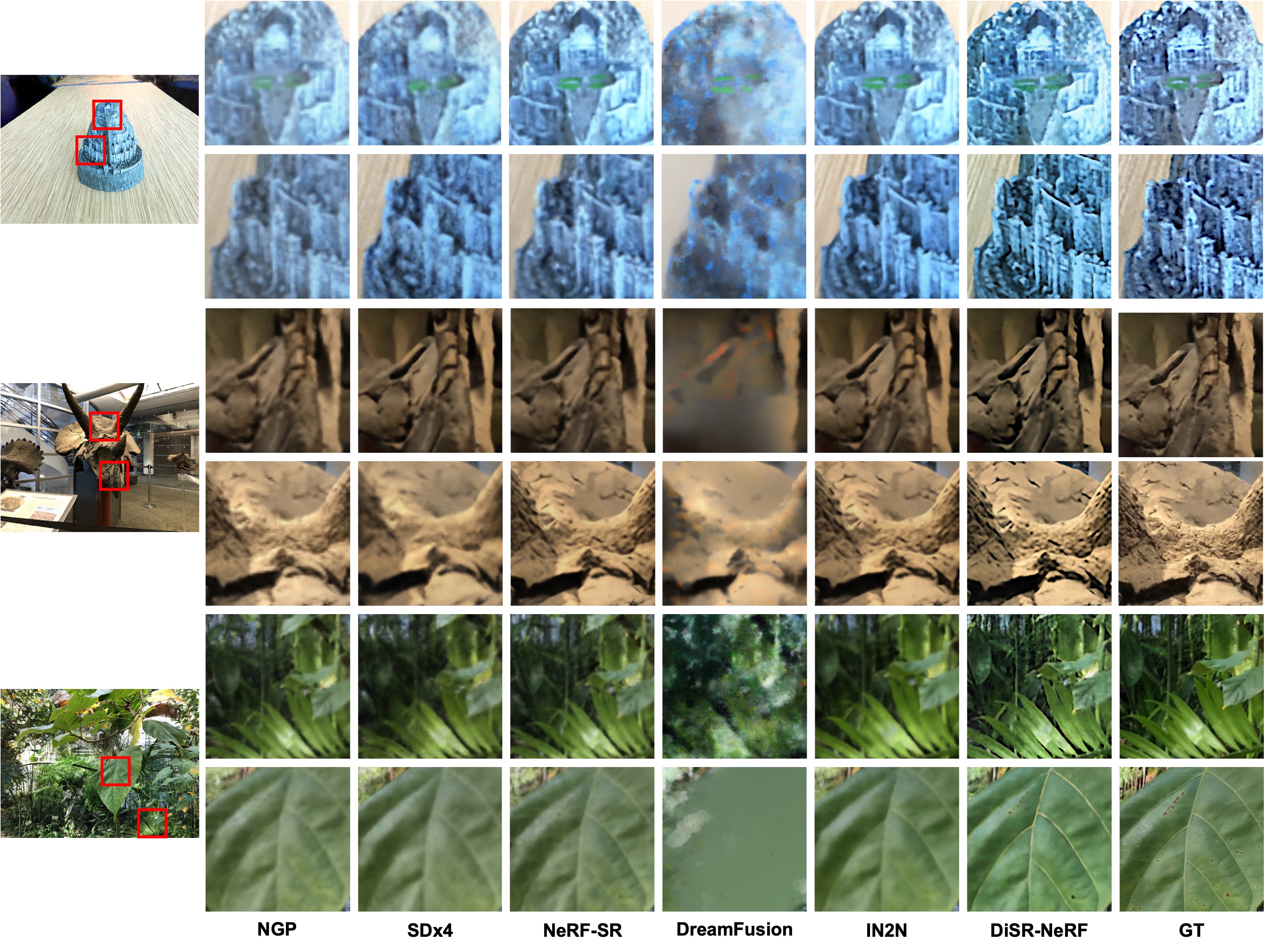}

\end{center}
\caption{
Qualitative Results on LLFF Dataset.}
\vspace{0 cm}
\label{fig:llff}
\end{figure*}
\section{Experiments}
\label{sec:experiments}
In this section, we provide both qualitative and quantitative comparisons to demonstrate the effectiveness of the proposed DiSR-NeRF. We also show ablation results under with different upscaling and 3D synchronization methods.

\subsection{Experimental Settings}
We compare DiSR-NeRF with five baseline models, all of which utilize the Instant-NGP \cite{muller2022instant} backbone:
\begin{itemize}
    \item \textbf{NGP.} We train NGP only on LR images as a baseline NeRF and render at HR resolution following \cite{han2023supernerf}.
    \item \textbf{SD$\times$4.} NeRF trained over images independently upscaled (via ancestral sampling) by the SD$\times$4 upscaler.
    \item \textbf{NeRF-SR.} We compare with NeRF-SR \cite{wang2021nerf-sr} without its HR refinement module as it requires HR reference images.
    \item \textbf{DreamFusion.} We adapt the SD$\times$4 upscaler to provide SDS guidance under the DreamFusion \cite{poole2022dreamfusion} framework, where SDS gradients are backpropagated through rendered image patches to the NeRF parameters.
    \item \textbf{IN2N.} We adapt Instruct-NeRF2NeRF \cite{instructnerf2023}, and replace the InstructPix2Pix editor with the SD$\times$4 upscaler. IN2N applies Iterative Dataset Update with ancestral sampling to gradually replace training images with upscaled images in each iteration.
\end{itemize}

\subsection{Dataset and Metrics}
We evaluate our models over the NeRF-synthetic \cite{mildenhall2020nerf} dataset containing 8 synthetic subjects and the real-world LLFF \cite{mildenhall2019llff} dataset containing 8 real-world scenes. On the NeRF-synthetic dataset, we use LR images of 400$\times$400 resolution for training, and we render at 1600$\times$1600 resolution. On the LLFF dataset, we train over LR images of 504$\times$378 resolution, and render at 2016$\times$1512 resolution. When evaluating each scene, we use 100 test poses distributed in a circular arrangement with all cameras directed towards the scene center.

Since SR details are produced by a generative model, different SR results can be valid for the same LR NeRF. Consequently, we do not compare our SR results against the HR ground truth. Instead, we follow \cite{han2023supernerf} to use the Naturalness Image Quality Evaluator (NIQE) \cite{niqe} as a no-reference evaluator to assess the quality of the rendered SR views. The NIQE metric is a blind image quality analyzer that measures statistical deviations against the natural scene statistics of natural, undistorted images. 

Following \cite{huang2021learning}, we use the warped LPIPS metric to assess consistency across viewpoints. The warped LPIPS metric is given by:
\begin{equation}
E_{warp}(I_v,I_{v'})= \text{LPIPS}(M_{v,v'}\cdot I_v, W(I_v')),
\end{equation}
where $I_v$ and $I_v'$ are renders from nearby viewpoints $v$ and $v'$, $W(\cdot)$ is a warping function and $M_{v,v'}$ is a warping mask. The LPIPS \cite{zhang2018perceptual} score is then computed between the target and warped renders over the masked regions. We use the predicted depths to backproject pixels in $I_v'$ to a point cloud in world space, and apply a point cloud rasterizer to render the warped image $W(I_v')$ in viewpoint $v$. In our experiments, we select $v'$ as the 3rd nearest test pose from $v$.

\subsection{Qualitative Results} The qualitative comparisons are shown in \cref{fig:blender} and \cref{fig:llff}. Firstly, we observe that DiSR-NeRF produces clearer edges and sharper details compared to all baselines. For example, \cref{fig:blender} (3rd row), our DiSR-NeRF is able to generate highly intricate details of the netting on the mast of the ship. Secondly, as discussed in \cref{subsection:i3ds}, DreamFusion (4th column) fails to converge and introduces severe blurring to the rendered views. Instead, DiSR-NeRF resolves this effectively by segregating upscaling and NeRF fitting with I3DS. Lastly, we compare DiSR-NeRF against IN2N (5th column) which uses ancestral sampling for upscaling. Across both datasets, we see that our DiSR-NeRF consistently produces sharper and well-defined details over IN2N. This validates the effectiveness of our proposed RSD optimization over ancestral sampling.

\begin{table}[h]
\resizebox{\columnwidth}{!}{%
    \centering
    \begin{tabular}{@{}lllll@{}}
        \toprule
        \multirow{2}{*}[-0.5em]{Methods} & \multicolumn{2}{c}{NeRF-Synthetic} & \multicolumn{2}{c}{LLFF}\\
        \cmidrule(lr){2-5} \\
        \addlinespace[-10pt]

        {} & NIQE $\downarrow$ & LPIPS $\downarrow$ & NIQE $\downarrow$ & LPIPS $\downarrow$\\
        \midrule
        NGP & \multicolumn{1}{c}{9.776}  & \multicolumn{1}{c}{0.262} & \multicolumn{1}{c}{9.163} & \multicolumn{1}{c}{0.284}  \\
        SDx4 & \multicolumn{1}{c}{6.286}  & \multicolumn{1}{c}{0.189} & \multicolumn{1}{c}{7.461} & \multicolumn{1}{c}{0.201}  \\
        NeRF-SR \cite{wang2021nerf-sr} & \multicolumn{1}{c}{6.012}  & \multicolumn{1}{c}{0.158} & \multicolumn{1}{c}{7.038} & \multicolumn{1}{c}{0.156} \\
        DreamFusion \cite{poole2022dreamfusion} & \multicolumn{1}{c}{8.624} & \multicolumn{1}{c}{0.252}  & \multicolumn{1}{c}{8.857} & \multicolumn{1}{c}{0.247} \\
        IN2N \cite{instructnerf2023} & \multicolumn{1}{c}{5.847}  & \multicolumn{1}{c}{0.173} & \multicolumn{1}{c}{6.473} & \multicolumn{1}{c}{0.157} \\
        DiSR-NeRF (Ours) & \multicolumn{1}{c}{\textbf{5.386}} & \multicolumn{1}{c}{\textbf{0.144}} & \multicolumn{1}{c}{\textbf{5.544}} & \multicolumn{1}{c}{\textbf{0.141}} \\
        \bottomrule
    \end{tabular}
}
\caption{Quantitative comparison between DiSR-NeRF and the baselines on NIQE and warped LPIPS.}
\label{tab:results}
\end{table}

\subsection{Quantitative Results}
We show the quantitative results in \cref{tab:results}. DiSR-NeRF shows significantly improved NIQE scores compared to all baselines, indicating that DiSR-NeRF is able to synthesize views with greater perception quality including increased detail and sharpness. Furthermore, our DiSR-NeRF also achieves better warped LPIPS scores, validating the effectiveness of I3DS in converging towards view-consistent details. Across both synthetic and real datasets, our DiSR-NeRF is able to effectively generate view-consistent SR details, validating its applicability to real-world scenarios.

\begin{figure}[t]
\begin{center}
\centering

\includegraphics[trim=0 30 0 0, scale=0.6]{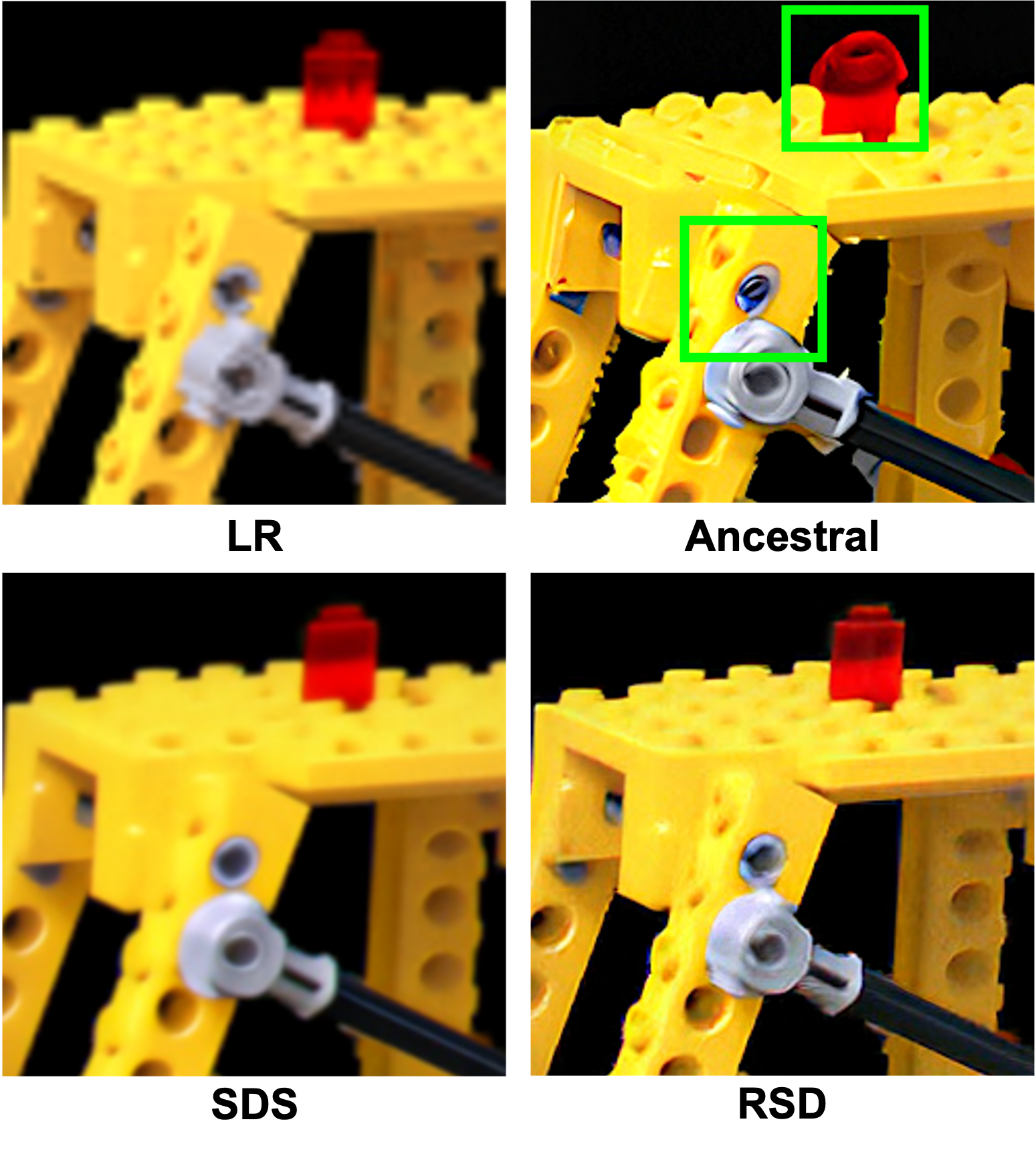}

\end{center}
\caption{
Comparison of 2D upscaling results. Green boxes highlight regions in ancestral sampling that deviate from LR conditioning (top left).}
\vspace{-0.cm}
\label{fig:2d}
\end{figure}
\begin{figure}[t]
\begin{center}
\centering

\includegraphics[width=\linewidth, trim=0 90 0 0]{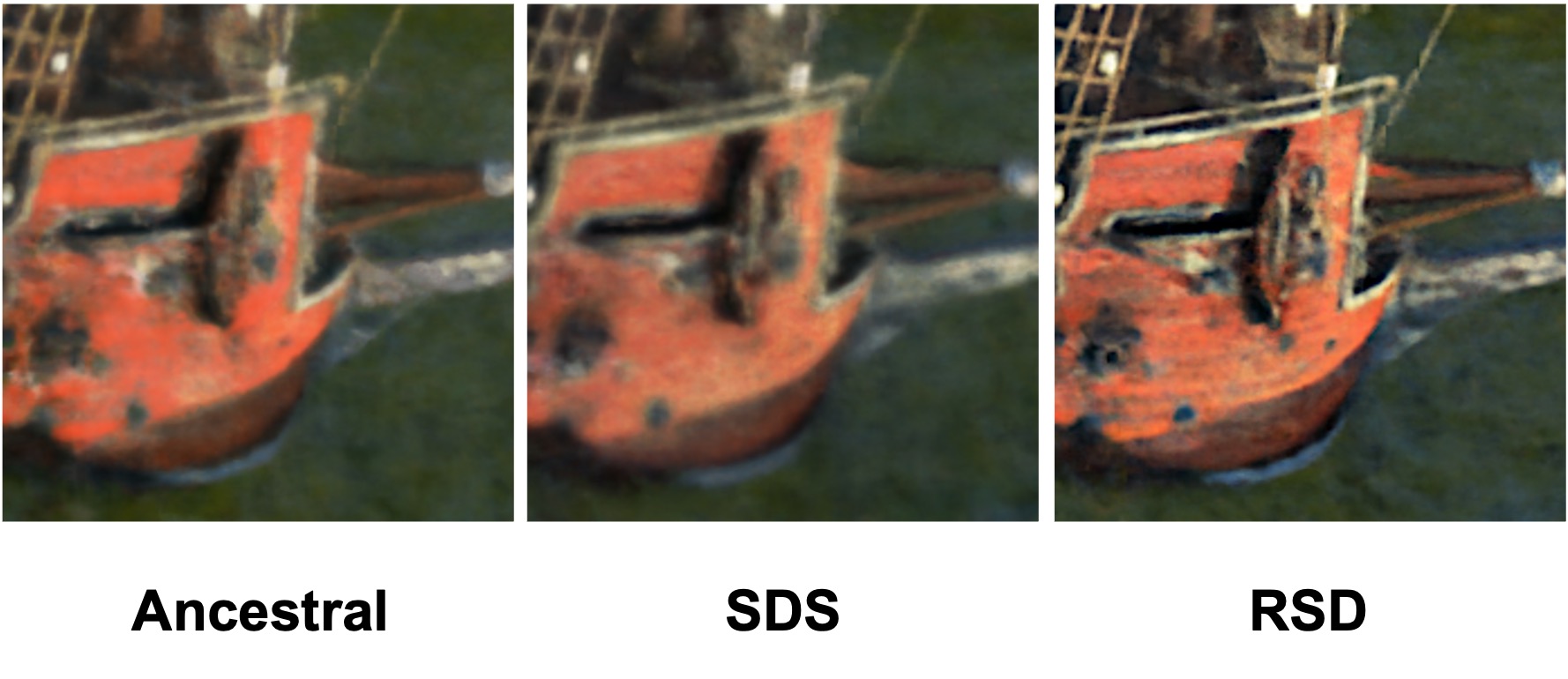}

\end{center}
\caption{
Quantitative ablations on upscaler method (Ancestral sampling, SDS RSD) on DiSR-NeRF.}
\vspace{-0.5cm}
\label{fig:ablations}
\end{figure}

\subsection{Ablations}
\label{subsection:ablations}
We also ablate the contributions of the various components in DiSR-NeRF. The quantitative results of the ablations are shown in \cref{tab:ablations} and the qualitative results in \cref{fig:ablations}. Firstly, we retain I3DS and replace the upscaling method with SDS and ancestral sampling instead of RSD (Row 1-2 in \cref{tab:ablations}). In both cases, replacing RSD results in an increase in NIQE and LPIPS scores, indicating poorer visual quality and lower view-consistency. This can also be observed in the visual ablations in \cref{fig:ablations}. Ancestral sampling fails to produce sharp details in the SR NeRF as it tends to generate high variance details which may be inconsistent to the LR image conditioning, as shown in our 2D experiments in \cref{fig:2d}. On the other hand, SDS produces NeRF renders with increased blur compared to RSD. This effect can also be observed from the 2D experiments in \cref{fig:2d}. We hypothesize that the mode-seeking property of the SDS objective optimizes images towards modes which may be far from typical samples \cite{nalisnick2019deep}. Thus, SDS may guide an image towards the mean of possible solutions which would result in blurred details. Unlike SDS, RSD follows an optimization trajectory similar to ancestral sampling, which guides an image towards more plausible samples.

\begin{table}[h]
\resizebox{\columnwidth}{!}{%
    \centering
    \begin{tabular}{@{}lllll@{}}
        \toprule
        \multirow{2}{*}[-0.5em]{Methods} & \multicolumn{2}{c}{NeRF-Synthetic} & \multicolumn{2}{c}{LLFF}\\
        \cmidrule(lr){2-5} \\
        \addlinespace[-10pt]

        {} & NIQE $\downarrow$ & LPIPS $\downarrow$ & NIQE $\downarrow$ & LPIPS $\downarrow$\\
        \midrule
        w/o RSD (SDS) & \multicolumn{1}{c}{6.273}  & \multicolumn{1}{c}{0.175} & \multicolumn{1}{c}{5.903} & \multicolumn{1}{c}{0.164}  \\
        w/o RSD (Anc.) & \multicolumn{1}{c}{5.942}  & \multicolumn{1}{c}{0.189} & \multicolumn{1}{c}{6.325} & \multicolumn{1}{c}{0.159}  \\
        w/o I3DS & \multicolumn{1}{c}{8.212}  & \multicolumn{1}{c}{0.254} & \multicolumn{1}{c}{8.299} & \multicolumn{1}{c}{0.236} \\
        DiSR-NeRF & \multicolumn{1}{c}{\textbf{5.386}} & \multicolumn{1}{c}{\textbf{0.144}} & \multicolumn{1}{c}{\textbf{5.544}} & \multicolumn{1}{c}{\textbf{0.141}} \\
        \bottomrule
    \end{tabular}
}
\caption{Ablations on I3DS and RSD in DiSR-NeRF.}
\label{tab:ablations}
\end{table}

We also ablate I3DS by replacing it with the DreamFusion framework. The results are shown in the 3rd row of \cref{tab:ablations}, which shows poor performance compared to our DiSR-NeRF. This indicates that I3DS is an essential component for super-resolution NeRF. Unlike Text-to-3D models which typically provide full image SDS supervision, the SR scenario requires high-resolution SDS guidance. This means small local patches need to be rendered at high resolutions, and the NeRF can only be supervised over a small region in each training step. As a result, each training step guides the NeRF in different optimization directions and thus preventing NeRF from converging towards high quality details. The segregation of the upscaling and NeRF synchronization processes in our I3DS allows RSD guidance to be efficiently applied without the memory constraints of online rendering. Moreover, I3DS also allows the NeRF synchronization to utilize batches of randomly sampled rays across all training views which allows for more stable convergence.

\begin{figure}[t]
\begin{center}
\centering

\includegraphics[width=\linewidth, trim=0 70 0 0]{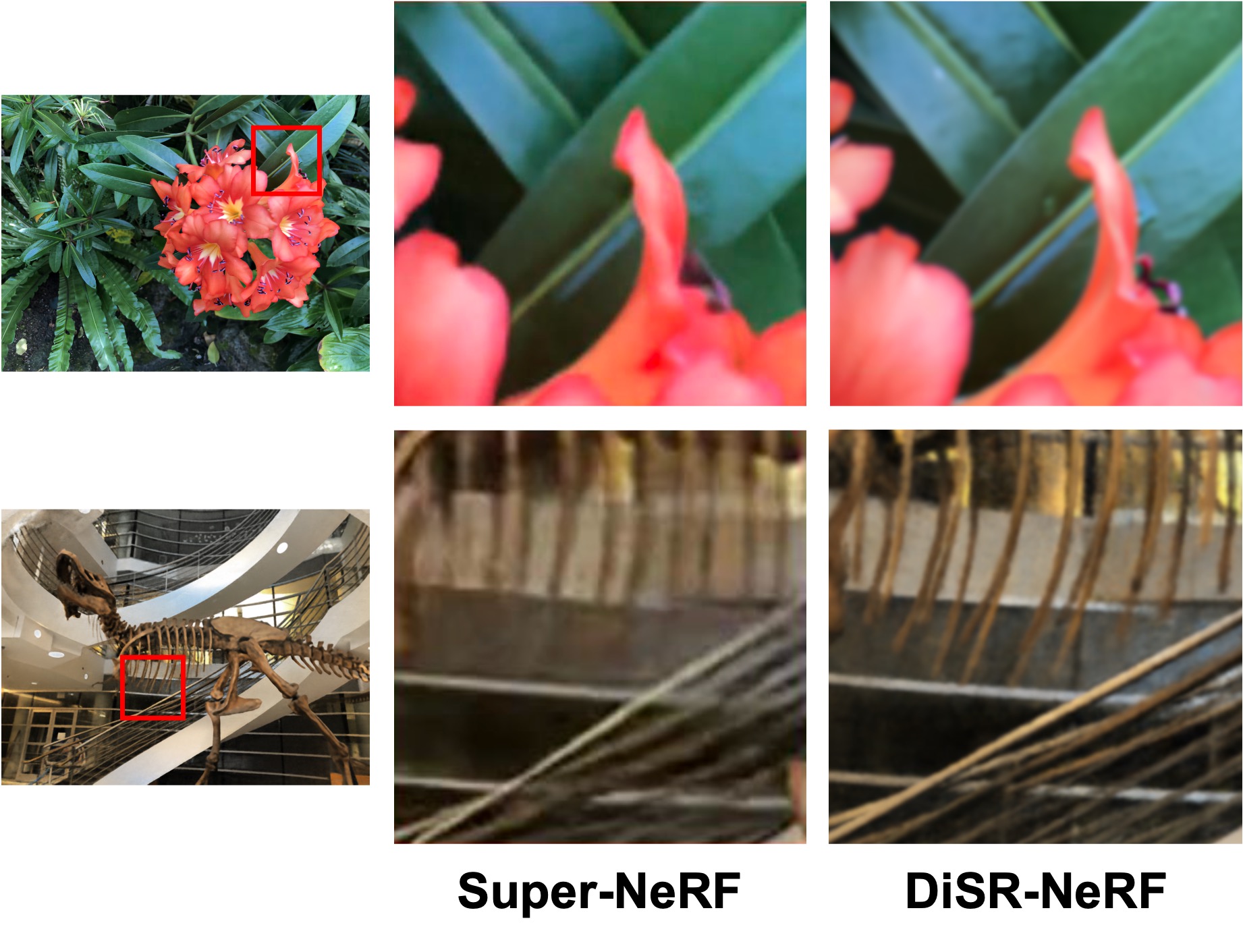}

\end{center}
\caption{
Qualitative comparison with Super-NeRF}
\vspace{-0.25cm}
\label{fig:super}
\end{figure}
\subsection{Comparison with Super-NeRF}
As the source code for Super-NeRF\cite{han2023supernerf} is not publicly available and the manuscript contains insufficient details on the evaluation method, we are unable to provide quantitative comparisons with the published results. Nonetheless, we conduct a qualitative assessment comparing Super-NeRF and DiSR-NeRF based on available visual results in the paper. The qualitative comparison is shown in \cref{fig:super}. Under similar settings, we observe that DiSR-NeRF can produce sharper details and clearer edges in both examples. In the LLFF-Flower scene, DiSR-NeRF is able to produce realistic textures on the leaves in the background and is able to resolve details on the stigma of the flower. In the LLFF-Trex scene, DISR-NeRF can generate clearer bone structures.
\section{Conclusion}
\label{sec:conclusion}
In conclusion, we propose DiSR-NeRF, a diffusion-guided super-resolution NeRF framework that distils 2D super resolution priors to the 3D domain to generate view-consistent high resolution details. 
DiSR-NeRF is able to achieve NeRF super-resolution without requiring high-resolution reference images or large multi-view image datasets. 
RSD achieves highly detailed, LR-consistent upscaling, while I3DS enables NeRFs to capture view-consistent details over successive upscaling and synchronization cycles.
We believe super-resolution methods such as DiSR-NeRF will have significant practical applications especially for devices equipped with low-resolution imaging capabilities.

\hfill \break
\textbf{Acknowledgements}. This research is supported by the National Research Foundation, Singapore under its AI Singapore Programmes (AISG Award No: AISG2-PhD/2021-08-012 and AISG Award No: AISG2-RP-2021-024).
{
    \small
    \bibliographystyle{ieeenat_fullname}
    \bibliography{main}
}

\setcounter{section}{0}
\renewcommand{\thesection}{\Alph{section}}
\clearpage
\setcounter{page}{1}
\maketitlesupplementary

\section{Limitations}
One limitation of DiSR-NeRF is the limited upscaling factor due to the use of the Stable Diffusion $\times$4 Upscaler which is designed for 4$\times$ super-resolution. In future work, we can consider applying cascaded diffusion models to achieve higher SR upscaling factors on low-resolution NeRFs.

\section{Pseudocode}\cref{alg:i3ds} and \cref{alg:rsd} shows the pseudocode for I3DS and RSD, respectively.

\begin{algorithm}
\caption{I3DS}
\label{alg:i3ds}
\textbf{Input}: LR NeRF $\omega_{lr}$, LR images $I_{lr}$, training poses $P_{tr}$

\textbf{Output}: SR NeRF $\omega_{sr}$
\begin{algorithmic}[1]
    \State $\omega = \omega_{lr}$
    \For {$\text{stage\_iter}=[0,\text{max\_stage\_iter}]$}

        \State //upscaling-stage
        \State $x_0=\Call{RenderImage}{\omega, P_{tr}}$
        \State $x_0 = \Call{InterpolateX4}{x_0}$
        \State $z_0=\Call{VaeEncode}{x_0}$ 
        \State $z'_0=\Call{RSD}{z_0,I_{lr}}$
        \State $x'_0=\Call{VaeDecode}{z'_0}$ 
        \State $I_{tr} = x'_0$

        \State //synchronization-stage
        \For {$\text{sync\_iter}=[0,\text{max\_sync\_iter}]$}
        \State $r_o,r_d,c_{tr}=\Call{SampleRays}{I_{tr},P_{tr}}$
        \State $\hat{c}=\Call{RenderRays}{r_o,r_d}$
        \State Take gradient descent step on $\nabla_\omega \|\hat{c}-c_{tr}\|$   
        \State $\omega_{old} \leftarrow \omega$
        \EndFor
    \EndFor
\State \Return $\omega_{sr} = \omega$
\end{algorithmic} 
\end{algorithm}

\begin{algorithm}
\caption{RSD}
\label{alg:rsd}
\textbf{Input}: Latent $z_0$, text prompt embeddings $y_{text}$, noise level $y_{noise\_level}$, min timestep $t_{min}$, max timestep $t_{max}$, LR images $I_{lr}$

\textbf{Output}: Refined latent residuals $h_\theta$
\begin{algorithmic}[1]
    \State $h_{\theta} = 0$ \Comment{Same shape as $z_0$}
    \For {$\text{sr\_iter}=[0,\text{max\_sr\_iter}]$}
        \State $\boldsymbol\epsilon\sim\mathcal{N}(0,\mathbf{I})$
        \State $y = y_{text}+y_{noise\_level}+I_{lr}$ 
        \State $t = t_{max} - (t_{max}-t_{min})\frac{\text{sr\_iter}}{\text{max\_sr\_iter}}$
        \State $z'_0 = z_0 + h_\theta$
        \State $z'_{t} = \sqrt{\bar{\alpha}_{t}}z'_0+\sqrt{1-\bar{\alpha}_{t}}\boldsymbol\epsilon$ \Comment{Eq. (1)}
        \State $z'_{t-1} =\sqrt{\bar{\alpha}_{t-1}}z'_0+\sqrt{1-\bar{\alpha}_{t-1}}\boldsymbol\epsilon$
        \Comment{Eq. (1)}
        \State$\boldsymbol\epsilon_\phi(z'_t,y,t)=\Call{Unet}{z'_{t},y,t}$
        \State $\hat{z}'_{t-1} = \frac{1}{\sqrt{\alpha_t}}\Bigl(z'_t-\frac{1-\alpha_t}{\sqrt{1-\bar{\alpha_t}}}\boldsymbol\epsilon_\phi(z'_t,y,t)\Bigr)+\sigma_t\boldsymbol\epsilon$ \Comment{Eq. (3)}
        \State Take gradient descent step on $\nabla_{\theta}\|z'_{t-1}-\hat{z}'_{t-1}\|$
        \State $h^{old}_\theta \leftarrow h_\theta$
    \EndFor
    \State \Return $h_\theta$

\end{algorithmic} 
\end{algorithm}

\section{Relating SDS to RSD}
We show the relation of our RSD to the existing SDS loss here.
As discussed in \cite{zhu2023hifa}, SDS can be reformulated into a difference of latent vector residuals $\mathbf{z}_0$ and $\hat{\mathbf{z}}_0$, where $\mathbf{z}_0$ is the latent vector under optimization and $\hat{\mathbf{z}}_0$ is the one-step denoised estimate of $\mathbf{z}$.

Starting with the SDS objective proposed in \cite{poole2022dreamfusion}:

$$\nabla_\theta \mathcal{L}_{\text{SDS}} = \gamma(t)(\hat{\boldsymbol\epsilon}_\phi(\mathbf{z}_t,y,t)-\boldsymbol\epsilon) \frac{\partial \mathbf{z}}{\partial\theta},$$
we substitute 
$$\hat{\boldsymbol\epsilon}_\phi(\mathbf{z}_t,y,t)=\frac{1}{\sqrt{1-\bar{\alpha}_t}}(\mathbf{z}_t-\sqrt{\bar{\alpha}_t}\hat{\mathbf{z}}_0)$$ 
from the reconstruction equation 
$$\hat{\mathbf{z}}_0=\frac{1}{\sqrt{\bar{\alpha}_t}}(\mathbf{z}_t-\sqrt{1-\bar{\alpha}_t}
\boldsymbol\epsilon_\phi(\mathbf{z}_t,y,t)),$$ 
and also substitute $\boldsymbol\epsilon=\frac{1}{\sqrt{1-\bar{\alpha}_t}}(\mathbf{z}_t-\sqrt{\bar{\alpha}_t}\mathbf{z}_0)$ from the forward noising process in Eq. (1). This gives us:
$$=\gamma(t)\Bigl(\frac{1}{\sqrt{1-\bar{\alpha}_t}}(\mathbf{z}_t-\sqrt{\bar{\alpha}_t}\hat{\mathbf{z}}_0)-\frac{1}{\sqrt{1-\bar{\alpha}_t}}(\mathbf{z}_t-\sqrt{\bar{\alpha}_t}\mathbf{z}_0)\Bigr)\frac{\partial \mathbf{z}}{\partial\theta},
$$
which reduces to:
$$=\gamma(t)\Bigl(\frac{1}{\sqrt{1-\bar{\alpha}_t}}\Bigr)\Bigl(-\sqrt{\bar{\alpha}_t}\hat{\mathbf{z}}_0+\sqrt{\bar{\alpha}_t}\mathbf{z}_0\Bigr)\frac{\partial \mathbf{z}}{\partial\theta},
$$
and finally
\begin{equation}
\nabla_\theta \mathcal{L}_{\text{SDS}}=\gamma(t)\Bigl(\frac{\sqrt{\bar{\alpha}_t}}{\sqrt{1-\bar{\alpha}_t}}\Bigr)\Bigl(\mathbf{z}_0-\hat{\mathbf{z}}_0\Bigr)\frac{\partial \mathbf{z}}{\partial\theta}.
\end{equation}
We can interpret this formulation of SDS as an optimization objective within the $\mathbf{z}_0$ space. With this formulation of SDS, RSD can be interpreted as a \textit{renoised} variant of SDS. Specifically, the conversion entails applying the forward noising process in Eq. (1) to both $\mathbf{z}_0$ and $\hat{\mathbf{z}_0}$ towards time $t-1$ which derives the RSD objective:
$$
\mathcal{L}_{RSD}=\|\mathbf{z}_{t-1}-\hat{\mathbf{z}}_{t-1}\|\frac{\partial{\mathbf{z}_{t-1}}}{\partial{\theta}}, 
$$
In practice, $\hat{\mathbf{z}}_{t-1}$ can be obtained directly from $\hat{\boldsymbol\epsilon}_\phi(\mathbf{z}_t,y,t)$  using the DDPM denoising equation in Eq. (3) instead of renoising $\hat{\mathbf{z}}_0$. 

\begin{figure}[t]
\begin{center}
\centering

\includegraphics[width=\linewidth, trim=20 30 0 0]{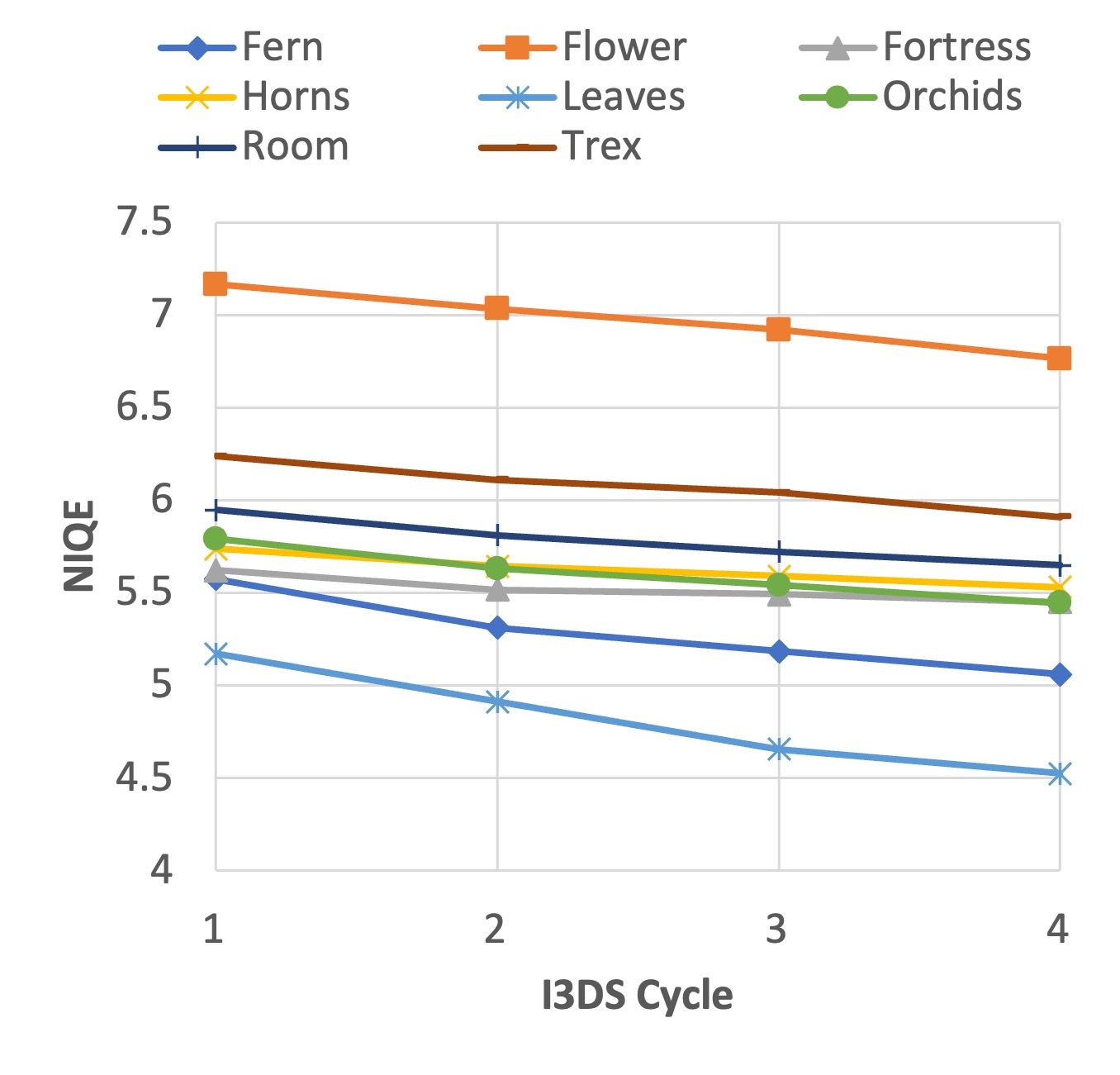}

\end{center}
\caption{
NIQE scores over successive I3DS cycles for scenes in LLFF dataset.
}
\label{fig:i3ds_conv}
\end{figure}
\section{I3DS Convergence}
In \cref{fig:i3ds_conv} we plot a graph of the NIQE scores DiSR-NeRF on LLFF scenes over successive I3DS cycles. Across all scenes, we see that NIQE improves with increasing I3DS cycles. This validates that the I3DS framework indeed enables NeRF $\boldsymbol\omega$ to converge onto high quality, view-consistent details.

\begin{table}[h]
\scriptsize
\resizebox{\columnwidth}{!}{%
    \centering
    \begin{tabular}{@{}lllll@{}}
        \toprule
        \multirow{2}{*}[-0.5em]{Methods} & \multicolumn{2}{c}{NeRF-Synthetic} & \multicolumn{2}{c}{LLFF}\\
        \cmidrule(lr){2-5} \\
        \addlinespace[-10pt]

        {} & PSNR $\uparrow$ & SSIM $\uparrow$ & PSNR $\uparrow$ & SSIM $\uparrow$\\
        \midrule
        NGP & \multicolumn{1}{c}{31.95}  & \multicolumn{1}{c}{0.959} & \multicolumn{1}{c}{18.02} & \multicolumn{1}{c}{0.617}  \\
        SDx4 & \multicolumn{1}{c}{28.94}  & \multicolumn{1}{c}{0.935} & \multicolumn{1}{c}{14.31} & \multicolumn{1}{c}{0.519}  \\
        NeRF-SR \cite{wang2021nerf-sr} & \multicolumn{1}{c}{\textbf{32.86}}  & \multicolumn{1}{c}{\textbf{0.962}} & \multicolumn{1}{c}{\textbf{18.27}} & \multicolumn{1}{c}{\textbf{0.633}} \\
        DreamFusion \cite{poole2022dreamfusion} & \multicolumn{1}{c}{26.21} & \multicolumn{1}{c}{0.920}  & \multicolumn{1}{c}{14.11} & \multicolumn{1}{c}{0.502} \\
        IN2N \cite{instructnerf2023} & \multicolumn{1}{c}{28.62}  & \multicolumn{1}{c}{0.936} & \multicolumn{1}{c}{14.49} & \multicolumn{1}{c}{0.521} \\
        DiSR-NeRF (Ours) & \multicolumn{1}{c}{31.05} & \multicolumn{1}{c}{0.948} & \multicolumn{1}{c}{17.01} & \multicolumn{1}{c}{0.580} \\
        \bottomrule
    \end{tabular}
}
\caption{PSNR and SSIM scores between DiSR-NeRF and the baselines.}
\label{tab:rebuttal}
\end{table}
\section{PSNR \& SSIM Scores}
We report PSNR \& SSIM scores in \cref{tab:rebuttal}. DiSR-NeRF still achieves highest similarity scores among prior-based methods (DiSR-NeRF, IN2N, DreamFusion-SDS, SD$\times$4).

\begin{table}[h]
\resizebox{\columnwidth}{!}{
    \centering
    \begin{tabular}{@{}llllll@{}}
        \toprule
        \multicolumn{6}{c}{Optimization Time (Hrs)$\downarrow$}\\
        \midrule\\
        \addlinespace[-10pt]
        NGP & SD$\times$4 & NeRF-SR\cite{wang2021nerf-sr} & DreamFusion\cite{poole2022dreamfusion} & IN2N\cite{instructnerf2023} & DiSR-NeRF (Ours)\\
        \midrule
        \textbf{0.25} & 0.35 & 0.30 & 8.00 & 5.00 & 6.00\\
        \bottomrule
    \end{tabular}
}
\caption{Comparison of optimization times.}
\label{tab:opt_time}
\end{table}
\section{Optimization Time}
We show optimization times in \cref{tab:opt_time}. DiSR-NeRF’s optimization time is longer due to the RSD optimization, but is still faster than DreamFusion-SDS due to I3DS's segregation of upscaling and synchronization stages.

\section{Implementation Details}
\paragraph{NeRF Backbone.} In our DiSR-NeRF implementation, we use Instant-NGP \cite{muller2022instant} as our default NeRF backbone due to its fast training and rendering speed. 
We also utilize dynamic ray sampling to increase ray count when the occupancy grid is sufficiently pruned. This optimizes GPU memory usage for faster convergence. 

\vspace{-3mm}
\paragraph{Patch Sampling.} In our RSD upscaling stage, we sample uniform random crops of 128$\times$128 resolution in latent space, which corresponds to an image patch of 512$\times$512 resolution. Compared to full image optimization, we find empirically that patch cropping at 128$\times$128 resolution offers the fastest optimization speed and best upscaling performance across all scenes. 

\vspace{-3mm}
\paragraph{Text Prompt.} For the text conditioning, we use a fixed text prompt \textit{"$\langle$subject$\rangle$, high resolution, 4K, photo"} for all scenes and all patches, where the subject tag is replaced with the scene name defined in the dataset. 

\vspace{-3mm}
\paragraph{Learning Rate.} We use a constant learning rate of $1e-2$ for all learnable parameters. 

\vspace{-3mm}
\paragraph{I3DS.} In the I3DS training regime, we use 5000 upscaling steps with a batch size of 16 patches, followed by 20,000 NeRF training steps. We repeat this two stage cycle for 4 iterations in total.

% {
%     \small
%     \bibliographystyle{ieeenat_fullname}
%     \bibliography{main}
% }
\end{document}